\documentclass[letterpaper]{article}

\usepackage{amsmath} 
\usepackage{natbib,alifeconf}  
\usepackage{url,hyperref,cleveref}
\usepackage{booktabs}
\usepackage{graphicx} 
\usepackage{caption}
\usepackage{balance}
\title{Integrating Sample Inheritance into Bayesian Optimization for Evolutionary Robotics}

\author{
    K. Ege de Bruin$^{1}$,
    Kyrre Glette$^{1,2}$, \and
    Kai Olav Ellefsen$^{1}$ \\
    \mbox{}\\
    $^1$Department of Informatics, University of Oslo, Norway \\
    $^2$RITMO, University of Oslo, Norway\\
    egedebruin@gmail.com
}

\begin{document}

\maketitle

\begin{abstract}
In evolutionary robotics, robot morphologies are designed automatically using evolutionary algorithms. This creates a body-brain optimization problem, where both morphology and control must be optimized together. A common approach is to include controller optimization for each morphology, but starting from scratch for every new body may require a high controller learning budget. We address this by using Bayesian optimization for controller optimization, exploiting its sample efficiency and strong exploration capabilities, and using sample inheritance as a form of Lamarckian inheritance. Under a deliberately low controller learning budget for each morphology, we investigate two types of sample inheritance: (1) transferring all the parent's samples to the offspring to be used as prior without evaluating them, and (2) reevaluating the parent’s best samples on the offspring. Both are compared to a baseline without inheritance. Our results show that reevaluation performs best, with prior-based inheritance also outperforming no inheritance. Analysis reveals that while the learning budget is too low for a single morphology, generational inheritance compensates for this by accumulating learned adaptations across generations. Furthermore, inheritance mainly benefits offspring morphologies that are similar to their parents. Finally, we demonstrate the critical role of the environment, with more challenging environments resulting in more stable walking gaits. Our findings highlight that inheritance mechanisms can boost performance in evolutionary robotics without needing large learning budgets, offering an efficient path toward more capable robot design.
\end{abstract}

Data/Code available at: \url{https://tinyurl.com/alife2025-egedebruin}

\section{Introduction}
Traditionally, robots are designed manually by engineers to perform specific tasks. While methods like reinforcement learning are often used to optimize control systems, the morphology of most modern robots is still largely human-designed. This approach can result in efficient robots for predefined tasks. However, it limits the robot's ability to adapt to new tasks or environments, as its morphology is not optimized for adaptability beyond its initial design.

\begin{figure}[t!]
    \centering
    \includegraphics[width=0.7\linewidth]{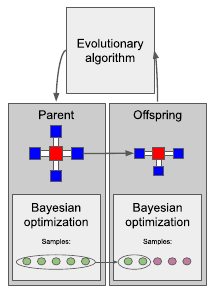}
    \caption{
        Overview of an evolutionary loop for morphology optimization with Bayesian optimization-driven controller learning, and controller inheritance methods.
    }
    \label{figure-evo-learn}
\end{figure}

In nature, we see many organisms that can quickly adapt to new environments and perform various tasks. This is the inspiration for evolutionary robotics, where robots are being designed using evolutionary algorithms \citep{Lipson2000, Nolfi2016, Sims1994}. A somewhat unique problem for robots is the co-optimisation of control and morphology \citep{Cheney2016, Faina2013}. The evolutionary algorithm needs to traverse a more complex search space due to the intertwined body and brain. Consequently, there might be a mismatch between control and robot morphology, and a robot's morphology might be discarded because it has not been performing to its full potential \citep{Eiben2013}. This causes the algorithm to stagnate quickly, without exploring much of the design space, because new, potentially well-performing robot morphologies are being discarded.

A common approach to this problem is to make a robot adapt its control to its new morphology, increasing the chance of performing to its morphology's potential. This is often done by adding an inner learning loop to the evolutionary algorithm: An outer evolutionary loop searches through the morphology search space. In contrast, an inner learning loop tries to find optimal control parameters for every new robot. Adding a learning loop is an efficient way to find well-performing robots \citep{Gupta2021, Luo2022, Miras2020, Zhao2020}. 

If every new robot morphology goes through a controller-learning phase, one could follow a Darwinian approach, where control parameters are either randomly initialized for every robot morphology \citep{Gupta2021}, or offspring robots inherit \emph{initial} control parameters from parents \citep{Miras2020}. It is also possible to follow a Lamarckian approach, where offspring robots inherit the \emph{learned} controllers of their parents \citep{Harada2024, Jelisavcic2019, Luo2023}. This way, information learned by parents can be transferred to offspring robots to kickstart the learning process.

A common problem in using Lamarckian inheritance is that offspring robots' morphologies are different from each other, where the number and location of sensors and actuators can differ. This raises the challenge of which information to transfer from parent to offspring. \cite{Jelisavcic2019} solve the problem by using a neural network as a genotype for both the evolutionary and learning processes, while other work keeps an archive of similar robots to inherit from \citep{LeGoff2022}, saves information for all possible locations in which an actuator can be located at \citep{Luo2023} or only transfers weights that are shared among neural network controllers \citep{Harada2024}.

Bayesian optimization is a learning algorithm with a fast learning capability and sample efficiency \citep{Chatzilygeroudis2019, Lan2021, LeGoff2022, VanDiggelen2024}. In this paper, we investigate methods to inherit information between generations in an evolutionary robotics setting using Bayesian optimization. Because we have to transfer information in a Bayesian optimization setting, we require different approaches than those used in other learning settings. Our main contributions are the following:

\begin{itemize}
    \item New techniques for inheriting control parameters in a Bayesian optimization controller learning setting.
    \item Analysis of how the advantage of inherited control is affected by the similarity between parent and offspring, and how it depends on the robot's environment.
\end{itemize}

To our knowledge, this is the first Lamarckian Inheritance method developed for controllers learned with Bayesian Optimization. This allows the sample-efficient adaptations from BO to be accumulated across generations of evolution.

\section{Related work}
\subsection{Body-brain optimization}
In the introduction, we discussed the problem of body-brain optimization, where a mismatch between robot morphology and control can cause stagnation and lack of diversity \citep{Eiben2013}. A common solution to this problem is to add controller optimization for every robot morphology, often called a learning loop \citep{Gupta2021, Luo2022, Miras2020, Zhao2020}. Learning usually requires a high learning budget to give the robot enough time to adapt to the morphology. In our work, we limit this learning budget using Bayesian optimization. Other approaches include \textit{innovation protection} \citep{Cheney2018}. It is similar to controller optimization in that robot morphologies are evaluated on multiple controller parameters, but the focus is more on adding this to innovative morphologies. Another approach is to train a single controller on a diverse set of morphologies, to be used as a teacher for new robot morphologies \citep{Mertan2024}. Finally, to deal with a lack of diversity \textit{quality-diversity} algorithms such as novelty search with local competition \citep{Lehman2011} and MAP-Elites \citep{Mouret2015, Nordmoen2021} have been used to enforce both the quality and diversity of solutions. These approaches need an explicit diversity measure to steer the algorithm towards more diverse solutions, and choosing this measure poses another challenge in the design of the algorithm \citep{Norstein2022}.

\subsection{Bayesian optimization}
For optimizing robot controllers, we apply Bayesian optimization \citep{Mockus1974}, which is known for its sample efficiency, and its ability to balance the exploration-exploitation trade-off. It uses a surrogate function to predict the objective value using the parameters to optimize, such as robot-control parameters in a robotics setting. Gaussian processes \citep{Williams2006} are commonly used as surrogate functions because they naturally handle uncertainty and can be customized to fit the problem setting \citep{Saito2024}. This makes the algorithm computationally expensive for high-dimensional problems with many samples, which is why other techniques are usually preferred for morphology optimization \citep{Bhatia2021}. However, with a limited dimensionality and sample size it is a good method for controller learning \citep{Lan2021, VanDiggelen2024}, where priors can also be exploited to kickstart the optimization process \citep{Chatzilygeroudis2019, Cully2015}

\begin{figure}[!t]
    \centering
    \includegraphics[width=0.95\linewidth]{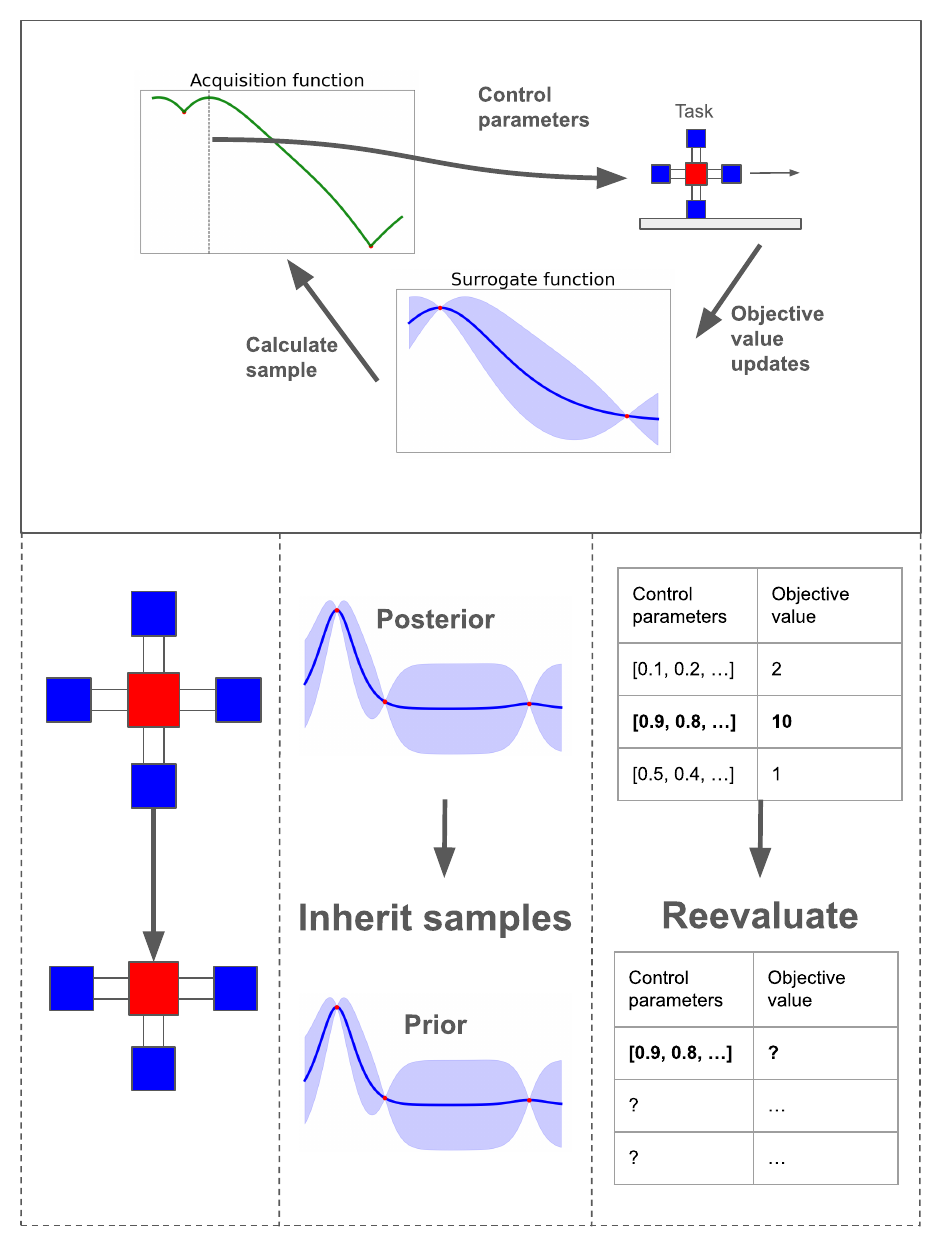}
    \caption{
        Overview of Bayesian optimization and the inheritance methods. Bayesian optimization iteratively samples the control parameters with the highest acquisition value, which considers the surrogate function's mean and uncertainty. Inheritance is done using two methods. With \textit{inherit samples}, all samples with their measured performance are transferred without evaluating them, this is the same as transferring the parent's posterior to be the offspring's prior. With \textit{reevaluate}, the best-performing samples from the parent are reevaluated by the offspring, leaving the rest of the learning budget for the Bayesian optimization process.
    }
    \label{figure-bo-inheritance}
\end{figure}

\subsection{Lamarckian evolution}
During the process of evolving robots, when new robots are created they often inherit control parameters from their parent robots. One could follow a Darwinian approach, where control parameters are either randomly initialized for every robot morphology \citep{Gupta2021}, or offspring robots inherit \emph{initial} (pre-learning) control parameters from parents \citep{Miras2020}. Another approach is to use Lamarckian evolution, where offspring robots inherit \emph{learned} control parameters from parents. Lamarckian evolution has received limited attention in evolutionary robotics, but studies have demonstrated its advantages over Darwinian evolution.  \cite{Jelisavcic2019} showed that inheriting learned controllers, rather than initial populations, improves performance. Similarly, \cite{Luo2023} found that passing learned parameters to offspring enhances adaptation. \cite{LeGoff2022} explored a related approach, using an archive to ensure inherited parameters remained compatible across different robot configurations. Lastly, \cite{Harada2024} use a transfer learning approach where neural network weights are shared, and they show an improvement when using crossover as well. All these works have shown that bootstrapping the learning process of robot morphologies using learned information from previous generations can improve the whole evolutionary process. In our work, we apply Lamarckian inheritance in a Bayesian optimization setting, showing the generalizability of Lamarckian inheritance in evolvable robots. We use a different form of Lamarckian inheritance closer to social learning, which allows an agent to learn by combining its experiences with experiences from other agents \citep{Bartoli2020}. Other works transfer control parameters directly, while we transfer samples of the Bayesian optimization process to either reevaluate them or use them as prior for the next agent.

\section{Sample inheritance}

Bayesian optimization samples parameters iteratively in the following manner: First, it uses a surrogate function to predict performance as a function of parameters. Then, it uses an acquisition function to sample the next most promising control parameters. It makes this prediction using the predicted mean and uncertainty for an optimal exploration-exploitation trade-off. The parameters are then evaluated, and the resulting objective value is used to update the surrogate. In our evolutionary robotics setting, the objective value is predicted as a function of the \textit{control parameters}. This prediction is based on previous samples. Therefore, with more samples, the prediction becomes more accurate.

In an evolutionary robotics setting, the controller optimization process can be initialized from scratch for every new morphology, and this is what others \citep{deBruin2025, Gupta2021, Luo2022, Miras2020, Zhao2020} have done in previous work. A common reason for starting from scratch is that there is a need for a mapping between parent and offspring to transfer information. This mapping can limit the choice of controllers. However, starting from scratch loses potential information that could be transferred from the parent robot to the offspring robot. This work investigates methods to transfer experience over generations in a setting with Bayesian optimization as controller optimization. We do this by transferring samples from parent robots to offspring robots. The sample inheritance approaches we try in this work are the following:

\begin{enumerate}
  \item Inherit samples: Offspring robots inherit samples from their parents, and use them as already sampled points, while not reevaluating them.
  \item Reevaluate: Offspring robots reevaluate the best samples from their parents.
\end{enumerate}

After inheritance, all robots continue sampling further using Bayesian optimization in both cases. In the \textit{inherit samples} approach, the offspring inherits all samples, including the measured performance of the parent robot, without reevaluating them. The upside of this approach is that more information is being transferred, even though the information is less accurate because it is not being evaluated on the same robot (since the parent likely had a different morphology than the offspring). In the \textit{reevaluate} approach, the offspring reevaluates the best samples from the parent. The upside of this approach is that the samples' measured performance will be more accurate because the control parameters are evaluated on the actual offspring robot. We ensure that the learning budget for both approaches is the same, meaning that when reevaluating the samples in the \textit{reevaluate} approach, these evaluations are taken from the offspring's budget. An overview of these approaches can be seen in Figure \ref{figure-bo-inheritance}.

\section{Method}

\subsection{Robot phenotype}
\begin{figure}[t!]
    \centering
    \includegraphics[width=0.6\linewidth]{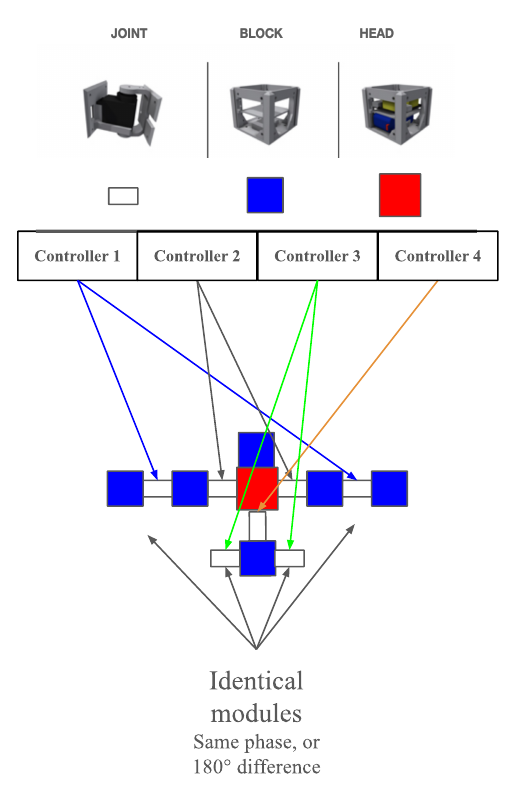}
    \caption{
        There are three types of modules, and every robot always has one head module. The head module has four attachment points, the block module has six (it includes up and down) and the joint module has two. Symmetry is ensured by making the left and right parts of the central axis identical. Every joint module has its own sine wave parameters, but the symmetrical parts share parameters.
    }
    \label{figure-overview}
\end{figure}

For this work, we make use of Revolve2\footnote{See https://github.com/ci-group/revolve2}{} as a modular robot framework, the modules are based on the RoboGen framework \citep{Robogen}. These modules are designed in the MuJoCo simulator, and it is possible to 3D print and attach them, to evaluate the robots in real life. The morphologies are built by attaching building blocks using three different modules: A head module, a block module, and a joint module. There is no restriction on the number of block and joint modules, but every robot has exactly one head module. The head module has four attachment points, the block has six, and the joint has two. Figure \ref{figure-overview} shows how the three modules look.

\begin{figure}
    \centering
    \includegraphics[width=0.5\linewidth]{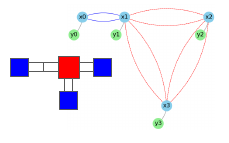}
    \caption{
        An example robot with its CPG network. The white modules are the actuators. A blue solid line represents a Manhattan distance of 1, and a red dotted line represents a Manhattan distance of 2.
    }
    \label{figure-cpg}
\end{figure}

The joint module is the only movable module. We use a decentralized controller and use two controller mechanisms to actuate the joints. Both mechanisms offer a simple and effective approach to generating rhythmic patterns for locomotion. The first is a sine-wave controller using the following equations:

\begin{equation}
\begin{split}
\Theta & = A * sin(\phi + P) + O \\ \phi & = \phi + \Delta_\phi * F
\end{split}
\end{equation}

In these equations, $\Theta$ is the output for the controller, $A$ is the amplitude, $P$ is the phase offset, $O$ is the angular offset, and $F$ is the frequency. $A$, $P$ and $O$ are the three learnable parameters, and $F$ is set to 4.

The second controller mechanism is a Central Pattern Generator (CPG) controller. CPGs are neural networks responsible for animals' rhythmic movement, without sensory information or rhythmic input \citep{Ijspeert2008}. The idea of the CPG network used in our network is to have neurons that work as differential oscillators, creating more complex oscillating patterns than a simple sine wave. In our work, each joint $i$ has two neurons, $x_i$ and $y_i$, that are connected using the following equations, where the $x_i$ neuron is responsible for the actuator's output. Note that $N_i$ is the set of neighboring joints of joint $i$, a neighbor is any joint within a Manhattan distance of 2.

\begin{equation}
    \Delta x_i = w_iy + \sum_{j \in N_i}w_{ji}x_j
\end{equation}

\begin{equation}
    \Delta y_i = -w_ix
\end{equation}

The initial state of each neuron is $\frac{1}{2}\sqrt{2}$. We define three learnable parameters for each joint: $w_i$ for the internal connection, $w_{ji_1}$ for each neighbouring joint with a Manhattan distance of 1, and $w_{ji_2}$ for each neighbouring joint with a Manhattan distance of 2. Figure \ref{figure-cpg} shows an example robot with its CPG network.

Even though the control is decentralized for both the sine-wave and the CPG controller, joints may share parameters due to symmetry. This will be explained further in the next subsection. An overview of the robot phenotype can be seen in Figure \ref{figure-overview}.

\begin{figure}[t!]
    \centering
    \includegraphics[width=1.0\linewidth]{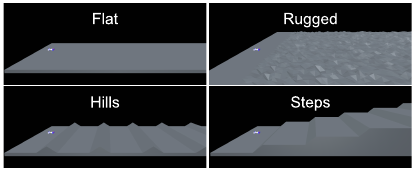}
    \caption{
        The four environments used in this work.
    }
    \label{figure-environments}
\end{figure}

\subsection{Robot genotype} \label{genotype}
The robot morphology is directly encoded in the genotype and can be represented as a tree. Every robot has a head module as the root of the three. We ensure symmetry by making the left and right sides of the robot identical. Therefore, there are three attachment points to the head module in the genotype, one for the front, one for the back, and one for the left and right sides of the head module. This is the same for all the block modules in the center of the robot. The head module is always in the centre of the robot, and every module placed in front or at the back of the core module is also labelled as a center module. Every joint module in the genotype has a sine wave as a controller. Due to symmetry, a joint on the left side of the centre has the same sine-wave parameters as its equivalent on the right side. We enforce symmetry to more closely resemble real-life creatures and to reduce the search space.

There are three possible \textit{body mutations}: Firstly, a random number of modules can be added to the robot. Secondly, a random number of modules can be removed from the robot. Lastly, there is a parameter that decides whether the symmetrical parts of the robot are in the same phase or in an alternating phase, and the third mutation switches this. We add the possibility of an alternating phase between the symmetrical parts to enforce a walking-like pattern. The third mutation only affects the sine-wave controller. For the CPG controller there is no specific parameter that can alternate the phase between symmetrical parts, which is why for the CPG controller this mutation has no effect.

In the previous section, it was shown that both the sine-wave and CPG controller have three parameters per joint. When a new joint is created, a new controller is made with random parameters, and when one is removed, this controller is also removed. For the evolutionary search of control parameters, we use a \textit{controller mutation} that changes these parameters by adding or subtracting Gaussian noise from these parameters. 

\begin{table}
\centering
\begin{tabular}{@{}lccc@{}}
\toprule
\textbf{Method} & \textbf{Inherited} & \textbf{Reevaluated} & \textbf{Fresh} \\
\midrule
Inherit Samples & 30 & 0 & 30 \\
Reevaluate      & 5  & 5 & 25 \\
\bottomrule
\end{tabular}
\caption{Number of samples \textbf{Inherited} from the parent, which of them are \textbf{Reevaluated}, and \textbf{Fresh} chosen by the Bayesian optimization process}
\end{table}

\subsection{Controller optimization}
The control learning algorithm is Bayesian optimization, which was chosen due to its fast learning capability and sample efficiency. Bayesian optimization has a high computational complexity, but due to our low number of controller parameters and low learning budget, this issue is minimal for our experiments. We use the Matern 5/2 kernel with a length scale of 0.2 as the surrogate function, and the Upper Confidence Bound as the acquisition function with an exploration variable of 3, which has been used and worked well before in evolving robots for directed locomotion \citep{Lan2021, VanDiggelen2024}. Samples evaluated by the robot have an uncertainty of 0. For the \textit{Inherit samples} experiments, we add an uncertainty of 2 to the inherited samples because this was the best-performing value in preliminary experiments.
We use a learning budget of 30 \cite{deBruin2025},
and in the \textit{reevaluate} setting, 5 samples are transferred from the parent robot to the offspring. This means that, with \textit{reevaluate}, offspring robots repeat the 5 best samples from their parents, leaving 25 samples for the Bayesian optimization process. In the \textit{inherit samples} setting, all 30 samples are transferred to the offspring robot without being evaluated, but only to be used as prior.

\begin{figure}[!t]
    \centering
    \includegraphics[width=1\linewidth]{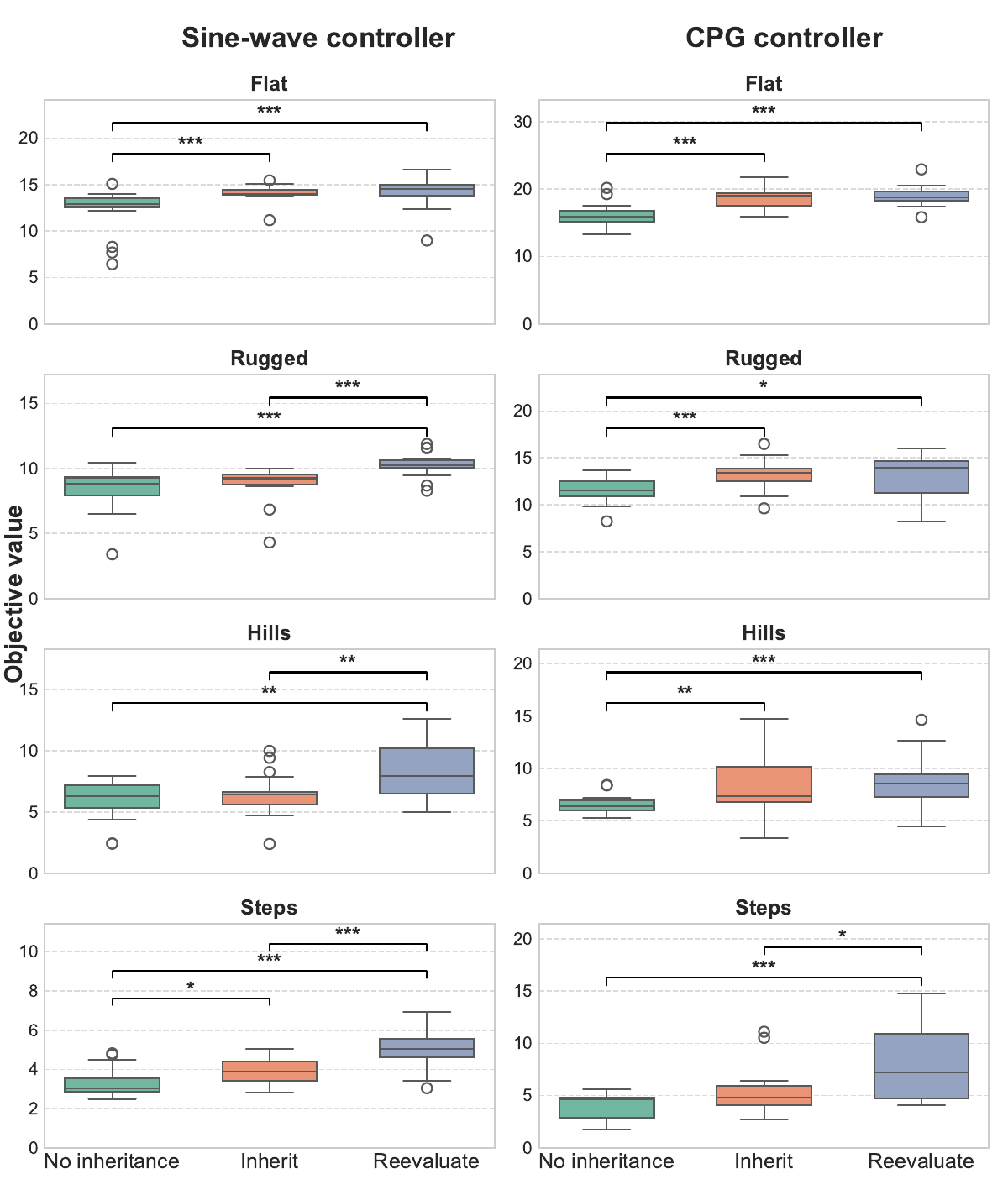}
    \caption{
        Boxplot of the best-performing robots for all 20 runs. We show significant differences where * is $p < 0.05$, ** is $p < 0.01$ and *** is $p < 0.001$.
    }
    \label{figure-performance}
\end{figure}

As explained above, inheriting controller samples is made challenging by the fact that the offspring may not have the same joints to control as its parent had. The direct encoding used in this work makes it possible to track which control parameters can be inherited. Two mutations can change the robot morphology: If a joint module is added, its control parameters are set to random, and if a joint module is removed, its control parameters are removed as well from the sample inheritance. The control parameters of the joints that have not been added or removed are transferred directly from the parent to the offspring robots.

\begin{figure}[t!]
    \centering
    \includegraphics[width=1.0\linewidth]{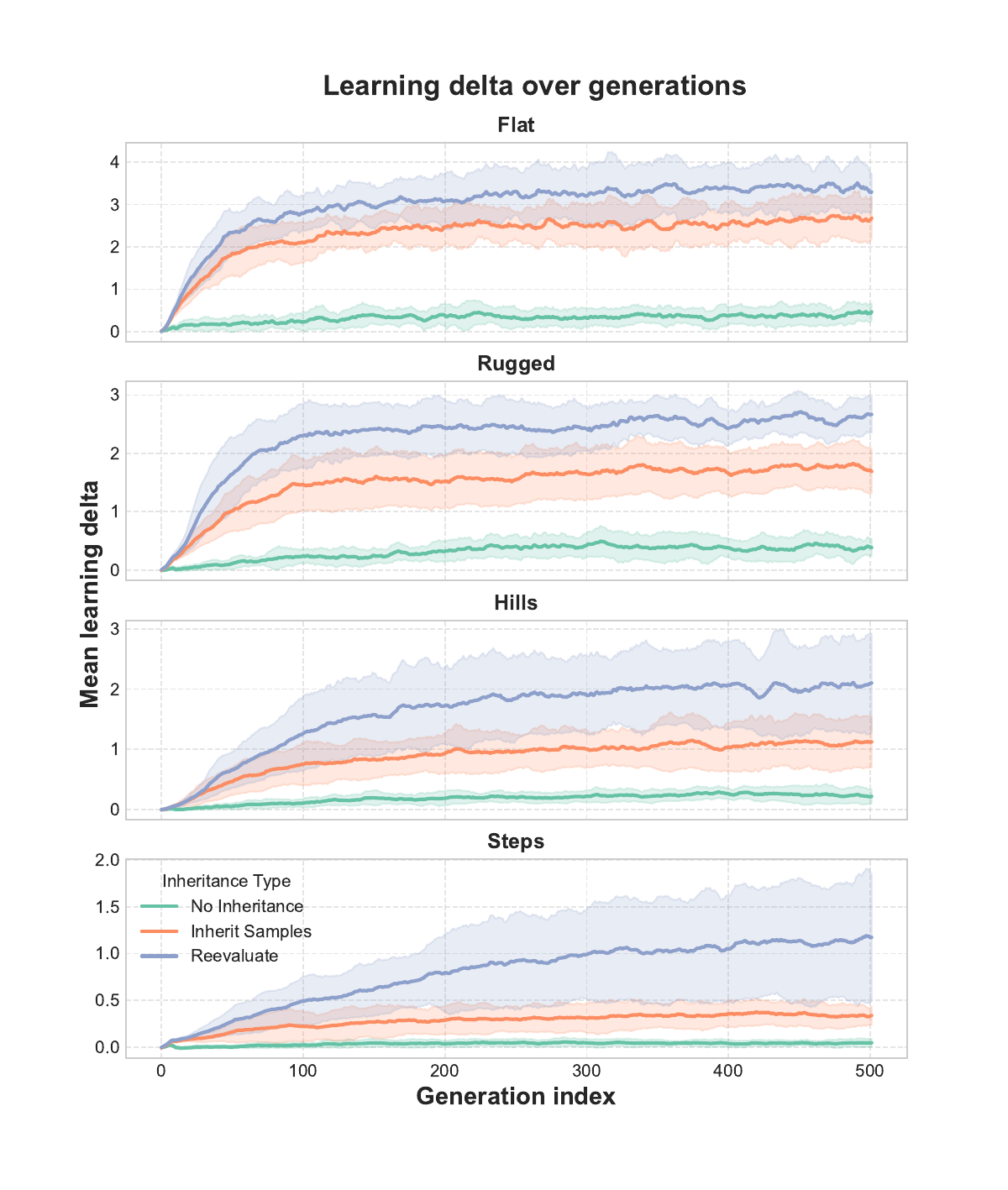}
    \caption{
    The learning delta is plotted over generations. We estimate the learning delta by calculating the advantage gained from the Bayesian optimization process. It is the robot's performance after learning compared to the robot with random control. We only show the learning delta for the sine-wave controller.
    }
    \label{figure-learn-delta}
\end{figure}

\subsection{Morphology optimization}
An evolutionary algorithm searches the robot morphology search space to find the best morphology for the task. The population size is 200, and every new robot morphology goes through a learning phase to find optimal control parameters. In each generation, 20 new robots are created by selecting parents through tournament selection with a tournament size of 5. The offspring replace the oldest robots in the population to ensure performance and diversity. New robots are created using asexual reproduction. We perform 500 generations, equalling around 300.000 function evaluations per run. Robots in the initial population have a size of between 15 and 20 modules, the add and remove mutation adds or removes up to 3 modules, and the maximum size is 20. If the robot size becomes too large, another mutation is tried. The size limit is chosen not to make the robots too large, as it should be possible to make them in real life. For the evolutionary search of the control parameters, Gaussian noise is added to the parameters with a mean of 0 and a standard deviation of 0.1.

There are four environments: Flat, hilly, rugged, and steps; see Figure \ref{figure-environments} for an example robot in the four environments. The task is to move forward as far as possible and the objective value of the robot is the displacement from beginning to end in 30 seconds. We compare the two sample-transferring approaches to a baseline where there is no sample transfer. In that case, the initial control parameters per morphology come from an evolutionary search with the mutation explained in the genotype section. For all methods, the morphology optimization is the same evolutionary algorithm as described. We perform 20 runs for each parameter configuration and perform Mann-Whitney U tests to test significant differences.

\section{Results}

\begin{figure}
    \centering
    \includegraphics[width=1.0\linewidth]{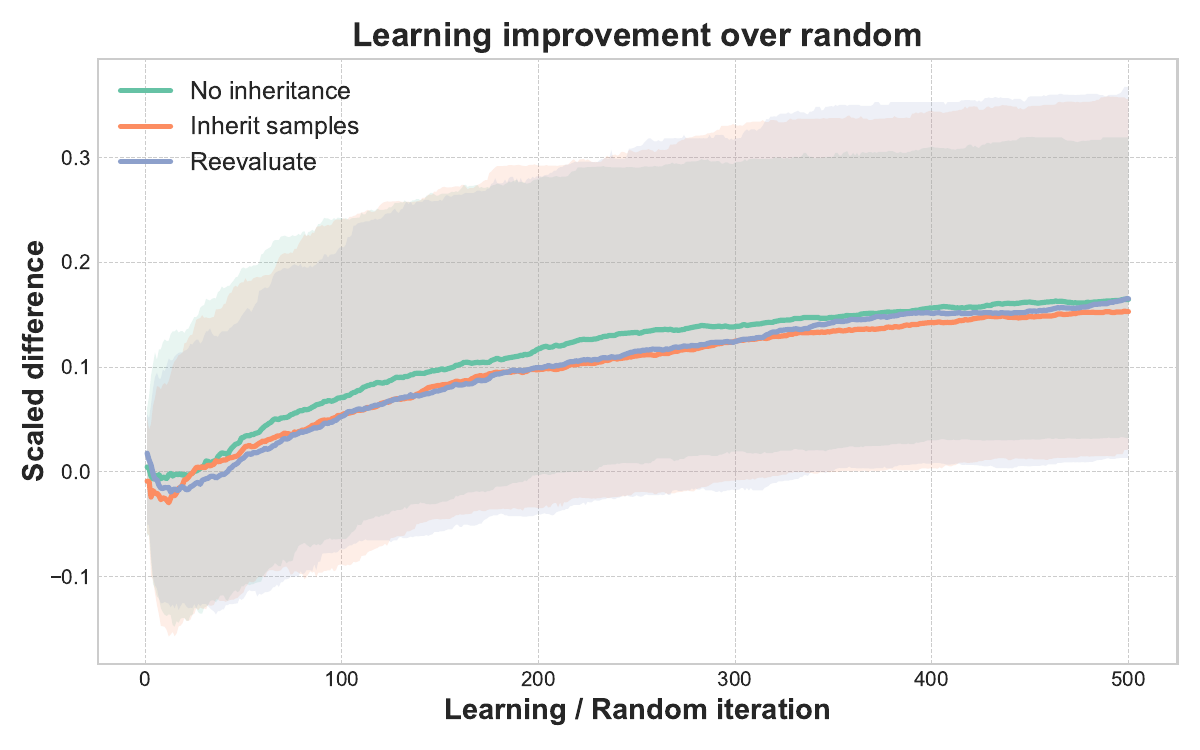}
    \caption{
       Average scaled improvement of learned over random controllers across iterations. Shaded areas show an interquartile range (25th to 75th percentile) to indicate variability. The analysis includes 20 randomly sampled robots per run, totalling 4.800 robots. All learning started from scratch; the colors of the lines show from which type of run the robots were taken.
    }
    \label{figure-learn-long}
\end{figure}

\subsection{Performance}
Figure \ref{figure-performance} shows the performance of the inheritance mechanisms with the sine-wave and CPG controllers, in the 4 different environments. For the sine-wave controller, \textit{reevaluate} consistently performs better, except for the \textit{flat} environment, where \textit{inherit samples} has similar performance. \textit{Inherit samples} performs better than \textit{no inheritance} only in the \textit{flat} and \textit{steps} environment, with a more significant difference in the \textit{flat} environment. For the CPG controller, both \textit{reevaluate} and \textit{inherit samples} perform better than \textit{no inheritance}, except for the \textit{steps} environment, where only \textit{reevaluate} has significantly better performance. Due to the limiting frequency of 4 for the sine-wave controller, we do not compare the two controllers.

Figure \ref{figure-learn-delta} shows the \emph{learning delta} over generations, which estimates how much the robot gains from Bayesian optimization, compared to a robot with a random controller. To avoid unfair disadvantages for random controllers, their performance is found by taking the best of 30 random samples on the given morphology. The learners spend these 30 samples searching for control parameters with Bayesian optimization, as normal. The results show that learning without Lamarckian inheritance (\textit{no inheritance}) does not perform much better than random sampling. However, adding inheritance does lead to an increase in the learning delta over generations. The main reason that \textit{no inheritance} does not improve much on a random search is likely due to the limited learning budget, and Figure \ref{figure-learn-long} supports this claim. We sampled a total of 4,800 robots from all runs and compared learning to random over a larger learning budget. The plot shows that with a larger learning budget, Bayesian optimization performs better than random - even without inheritance.

\subsection{Difference to parent}
\begin{figure}
    \centering
    \includegraphics[width=1\linewidth]{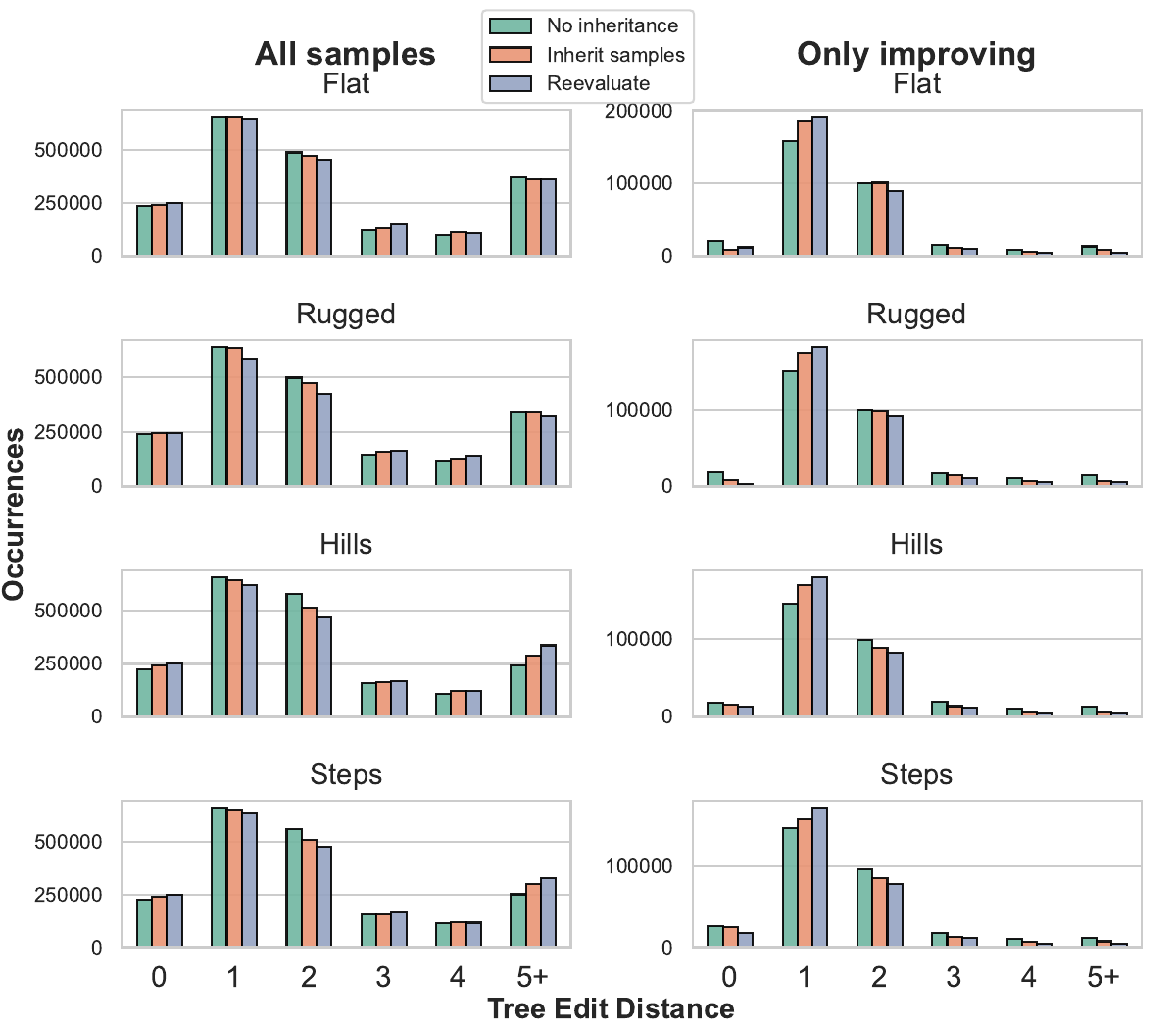}
    \caption{
        Bar plots showing the number of robots for the sine-wave controller where the tree edit distance is compared to its parent. The left plots show the occurrences for all robots and the right plot for those that have a higher fitness than their parent.
    }
    \label{figure-parent-ted}
\end{figure}
Not every mutation to a robot's morphology is successful, but a good learning algorithm should ensure that offspring have more successful mutations. In Figure \ref{figure-parent-ted}, we present the number of robots for each tree edit distance value, as used in previous work \citep{Samuelsen2014}, relative to their parent. We show it for all robots and for the robots that have a higher fitness than their parent. Most robots have a low tree edit distance to their parent, and with a tree edit distance of 1, there are more occurrences of \textit{improving} robots for \textit{reevaluate} and \textit{inherit samples} than \textit{no inheritance}.
This is despite that the two inheritance schemes have a fewer \textit{total} number of samples with a tree edit distance of 1 compared to \textit{no inheritance}.
Moreover, the number of improved samples for the inheritance schemes is already less than \textit{no inheritance} at a tree edit distance of 2, showing that inheritance is only beneficial for robots that are similar to each other. Finally, more robots are similar to their parents than those that are not similar to their parents. Note that the plot represents the sine wave controller, meaning that a robot with a tree edit distance of 0 from its parent differs in the phase between its symmetrical parts. In this case, the morphology likely requires a different control strategy.

\subsection{Environment}
\begin{figure}
    \centering
    \includegraphics[width=1\linewidth]{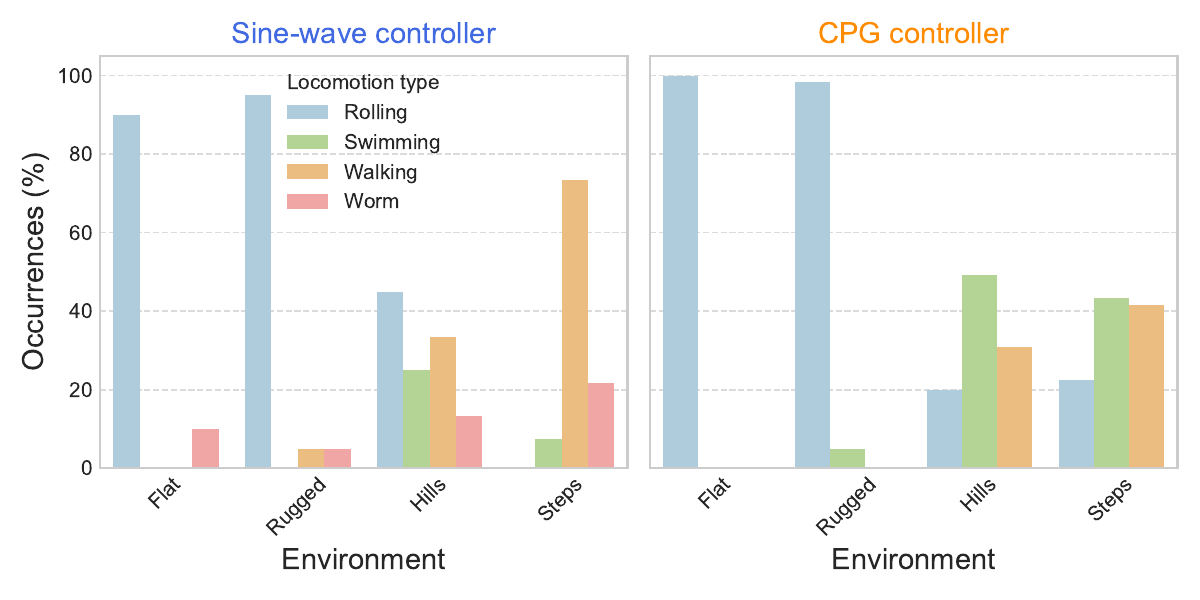}
    \caption{
        The locomotion type by environment of the best robot in every run. It shows the percentage of the locomotion type for every controller-environment combination. Swimming and walking are considered to be more stable than worming, which in turn is more stable than rolling.
    }
    \label{figure-move-type}
\end{figure}

\begin{figure}
    \centering
    \includegraphics[width=0.7\linewidth]{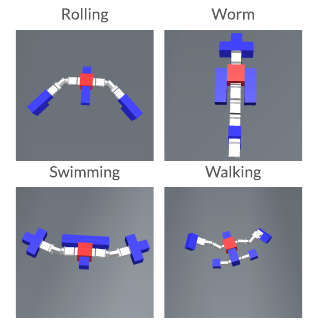}
    \caption{
        Typical robot morphologies for the four specified gaits.
    }
    \label{figure-robots}
\end{figure}

Figure \ref{figure-move-type} presents the distribution of locomotion types for the robot that performs the best in each run. We consider swimming and walking the most stable forms of locomotion, while rolling is the least stable. The results demonstrate the influence of the environment on locomotion strategy. Robots predominantly adopt rolling locomotion in both flat and rugged terrains. Environments with hills and steps favour more stable movement patterns, such as swimming and walking, showing more diversity in movement patterns. The main difference between controller mechanisms is the lack of the \textit{worm} locomotion type in the CPG controller. Figure \ref{figure-robots} shows typical robots for the four specified movement gaits\footnote{Videos of example robots can be found on the repository page: \url{https://tinyurl.com/alife2025-egedebruin}}{}.

\section{Discussion}
Our study of Lamarckian inheritance of robot control parameters using Bayesian optimization as controller optimization has indicated that inheriting samples works well. It works well both when reevaluating samples on the offspring, and when using parents' sampled values directly without reevaluation. We also show the importance of the environment, regardless of the inheritance mechanism. In more challenging environments, such as the hills and steps environment, the search resulted in more stable gaits.
Finally, we have shown that the method of transferring samples works with two types of controllers and four environments, where in more challenging environments it results in a more diverse set of walking gaits.

We also highlight the advantage of adding inheritance: it can improve performance without requiring a large learning budget. The findings suggest that incorporating inherited solutions into the offspring’s learning process can produce better results, even when the optimization algorithm itself does not perform particularly well or has a too-low budget. Additionally, the use of direct encoding with a low mutation rate supports the idea that similar parent robots lead to offspring with comparable or even improved performance.

It is important to consider how an indirect encoding might affect the results because the mapping from parent to offspring in such an encoding is not straightforward. An indirect encoding could potentially lead to greater diversity in offspring. This highlights an area for future work since exploring how this impacts the optimization process could show useful insights into the influence of Lamarckian inheritance.

\section{Conclusion}

We have investigated the influence of information exchange within generations of evolvable robots. To our knowledge, this is the first work to explore the combination of Bayesian optimization as a learning algorithm with Lamarckian inheritance. We transfer samples from parent to offspring in two methods: \textit{inherit samples} uses all parent samples as prior for the offspring, and \textit{reevaluate} uses the best samples from the parent robot and evaluates them on the offspring. We do this on a direct morphology encoding and use two decentralized controller mechanisms: a sine-wave controller and a two-neuron CPG controller. The results show that \textit{reevaluate} has the best performance and that \textit{inherit samples} often improves on \textit{no inheritance} as well.

Our robots are given a small learning budget, which lowers their benefit from learning in the standard case (no inheritance). However, adding an inheritance scheme greatly increases robots' \emph{learning delta}, suggesting that Lamarckian inheritance can be particularly valuable for Bayesian Optimization in settings where gathering new data is costly.

All methods work best with robots that are similar to their parents, and this difference is most prevalent for \textit{inherit samples} and \textit{reevaluate}. We therefore showed that our Lamarckian inheritance does not work well with robot bodies that are very different from each other. The evaluation of a robot for a task is the result of the interaction between its body and the environment. Therefore, future work can look into the effect of Lamarckian evolution on changing environments. Moreover, it would also be interesting to look into more complex tasks, environments, or controllers where controller optimization more important. 

\section*{Acknowledgements}
This work was partially supported by the Research Council of Norway through its Centres of Excellence scheme, project number 262762. The experiments were performed on the Fox Cluster, owned by the University of Oslo IT Department.

\footnotesize
\bibliographystyle{apalike}
\bibliography{references}

\end{document}